\documentclass[journal]{IEEEtran}

\usepackage[pdftex]{graphicx}
\graphicspath{{../pdf/}{../jpeg/}}
\DeclareGraphicsExtensions{.pdf,.jpeg,.png}

\usepackage{amsmath}
\usepackage{amssymb}

\usepackage[table,xcdraw]{xcolor}
\usepackage{array}
\usepackage{booktabs}
\usepackage{makecell}
\usepackage{multirow}
\usepackage{tabularx}
\usepackage{colortbl} 

\usepackage[ruled,linesnumbered]{algorithm2e}
\SetKwComment{Comment}{// }{}

\usepackage{cite}
\usepackage[numbers]{natbib}
\usepackage{url}

\usepackage{eqparbox}
\usepackage{xcolor,soul,framed}

\usepackage{hyperref}

\begin{document}
\bstctlcite{IEEEexample:BSTcontrol}
    \title{CoReVLA: A Dual-Stage End-to-End Autonomous Driving Framework for Long-Tail Scenarios via Collect-and-Refine}
  \author{Shiyu Fang,
      Yiming Cui,
      Haoyang, Liang,
      Chen Lv,
      Peng Hang,
      and Jian Sun
 \thanks{S. Fang is with the College of Transportation, Tongji University. (email: fangshiyu@tongji.edu.cn).}
}

\markboth{}{Roberg \MakeLowercase{\textit{et al.}}: High-Efficiency Diode and Transistor Rectifiers}

\maketitle

\begin{abstract}
Autonomous Driving (AD) systems have made notable progress, but their performance in long-tail, safety-critical scenarios remains limited. These rare cases contribute a disproportionate number of accidents. Vision-Language Action (VLA) models have strong reasoning abilities and offer a potential solution, but their effectiveness is limited by the lack of high-quality data and inefficient learning in such conditions. To address these challenges, we propose CoReVLA, a continual learning end-to-end autonomous driving framework that improves the performance in long-tail scenarios through a dual-stage process of data Collection and behavior Refinement. First, the model is jointly fine-tuned on a mixture of open-source driving QA datasets, allowing it to acquire a foundational understanding of driving scenarios. Next, CoReVLA is deployed within the Cave Automatic Virtual Environment (CAVE) simulation platform, where driver takeover data is collected from real-time interactions. Each takeover indicates a long-tail scenario that CoReVLA fails to handle reliably. Finally, the model is refined via Direct Preference Optimization (DPO), allowing it to learn directly from human preferences and thereby avoid reward hacking caused by manually designed rewards. Extensive open-loop and closed-loop experiments demonstrate that the proposed CoReVLA model can accurately perceive driving scenarios and make appropriate decisions. On the Bench2Drive benchmark, CoReVLA achieves a Driving Score (DS) of 72.18 and a Success Rate (SR) of 50\%, outperforming state-of-the-art methods by 7.96 DS and 15\% SR under long-tail, safety-critical scenarios. Furthermore, case studies demonstrate the model’s ability to continually improve its performance in similar failure-prone scenarios by leveraging past takeover experiences. All code, preprocessed datasets, and scenario configuration files are available at: \url{https://github.com/FanGShiYuu/CoReVLA}.

\end{abstract}


%
\IEEEpeerreviewmaketitle


\section{Introduction}





As Autonomous Driving (AD) technology continues to advance, Autonomous Vehicles (AVs) are gradually entering commercial deployment. McKinsey projects that by 2030, over 20\% of new vehicles will support Level 3 or higher automation \citep{mckinsey2023autonomous}. However, current systems still struggle in long-tail scenarios where driver intervention rates rise sharply \cite{lan2024hi, pan2024vlp}. This limitation points to a fundamental issue with modular autonomous driving systems: the accumulation of errors across perception, prediction, and planning stages makes further performance gains difficult \cite{le2022survey}. In contrast, end-to-end approaches that map sensor inputs directly to control actions offer greater adaptability and unified optimization \citep{chen2024end}. Therefore, these methods show promise in enhancing decision-making under long-tail scenarios, where conventional systems tend to falter and require human intervention \cite{hegde2025distilling}.

Current end-to-end methods can be broadly categorized into two types: small-scale task-specific models and large-scale pretrained models. Small-scale task-specific models typically process raw sensor inputs into structured intermediate representations such as BEV maps, actor-centric features, or interaction graphs \cite{chitta2022transfuser, gomes2023interaction, li2022time3d}. A unified model then jointly learns perception, prediction, and planning via multi-task objectives, enabling robust closed-loop performance \cite{jia2023think, zhang2025bridging}. By mitigating error accumulation and enabling closer subsystem integration, this paradigm has become a key direction in end-to-end autonomous driving. While task-specific networks are effective in most routine driving scenarios, their limited contextual reasoning and poor generalization to unseen situations hinder their performance in long-tail and complex environments.

On the other hand, pretrained models—especially Vision-Language Models (VLMs)—bring extensive world knowledge and strong reasoning abilities, making them a compelling alternative for autonomous driving tasks \cite{guo2024vlm}. Several studies have explored integrating VLMs into autonomous driving subtasks, such as scene comprehension \cite{zhang2024think}, anomaly detection \cite{wang2024drive}, and interaction \cite{xu2024drivegpt4, chen2025solve}. Consequently, when encountering complex scenes, VLMs emulate human-like reasoning, progressing from perception to interpretation to action, forming a Vision-Language-to-Action (VLA) framework. Recent studies suggest that the cognitive reasoning capabilities of VLA can improve decision-making in uncertain or high-stakes environments, thereby enabling more coherent and context-aware driving behaviors \cite{arai2025covla, jiang2024senna}. Therefore, VLA is considered a promising direction for enhancing autonomous driving performance in long-tail scenarios \cite{tian2024drivevlm}.

Despite their potential, VLAs still face significant challenges when deployed in long-tail scenarios, including: \textbf{1) Scarcity of Long-Tail QA Data:} Most public autonomous driving datasets focus on trajectory-level annotations and lack raw visual data. Moreover, existing QA datasets tailored for vision-language models rarely include long-tail scenarios. Therefore, acquiring high-quality long-tail scenario QA data has become a key common challenge across current research efforts. \textbf{2) Inefficient Fine-Tuning under Sparse Data:} Due to the inherently low frequency of long-tail scenarios, enabling the model to learn effectively from limited data has become another key challenge in improving its performance under such conditions.


To tackle the above challenges, we introduce a continual learning CoReVLA framework for end-to-end autonomous driving via a \textbf{Collect-and-Refine} dual process. First, multiple open-source driving QA datasets are aggregated to perform Supervised Fine-Tuning (SFT). Then, the model is deployed within the immersive Cave Automatic Virtual Environment (CAVE) Human-in-the-loop (HITL) testing platform, where its behavior and the corresponding driver takeovers are \textbf{collected} and reconstructed into QA data form. Finally, Direct Preference Optimization (DPO) is employed, leveraging human takeovers as preference feedback to \textbf{refine} the model’s behavior in long-tail scenarios. The contributions of our work are summarized as follows:

\begin{itemize}
\item \textbf{Collection of visually grounded takeover data via HITL testing in the immersive CAVE platform}. The CAVE platform reconstructs 3D scenarios from trajectories, enabling end-to-end AD testing. During testing, long-tail scenarios where the model underperforms are proactively taken over by human drivers, yielding valuable takeover data including visual context, driver behaviors, and real-time attention.

\item \textbf{Introduction of the DPO approach for efficient behavior refinement from sparse takeover data}. By contrasting suboptimal pre-intervention behaviors from models with high-quality human takeovers, the CoReVLA directly learns driver preferences, avoiding the pitfalls of indirect reward modeling and significantly improving learning efficiency.

\item \textbf{Validation of CoReVLA in both open-loop and closed-loop settings}, demonstrating effective scene understanding and decision-making capabilities. On the Bench2Drive benchmark, CoReVLA achieves a Driving Score of 72.18 and a Success Rate of 50\%, surpassing SOTA methods by 7.96 and 15\% respectively in long-tail, safety-critical scenarios. Case studies further verify its potential for cross-scenario generalization capability.
\end{itemize}








\section{Related Work}
\subsection{Small-scale task-specific models for AD}
Recent advances in AD have led to the emergence of unified frameworks that integrate perception, prediction, and planning into a single model \cite{singh2023end, chitta2021neat}. The key idea is to leverage multi-task learning to jointly model the spatial semantics of the environment, the motion patterns of surrounding agents, and the ego vehicle's decision-making, enabling fully differentiable optimization across the entire pipeline \cite{huang2024dtpp, xu2021end}. Compared to traditional modular systems, these models exhibit stronger temporal consistency and cross-task coordination \cite{chang2021yoltrack, ishihara2021multi}. However, due to their reliance on limited-scale datasets and fixed task priors, such frameworks tend to overfit to seen scenarios and struggle to generalize in long-tail or highly interactive situations, where diverse behaviors, occlusions, and ambiguous intentions pose substantial challenges \cite{zhang2023deep, girase2021loki, sima2024drivelm}.

\subsection{Large-scale pretrained models for AD}
On the other hand, an emerging line of research has explored the potential of large-scale pretrained models to enhance various aspects of autonomous driving \cite{NIE2025100003}. Large Language Models (LLMs), by virtue of their world knowledge and abstraction capabilities, have been introduced to support tasks such as decision explanation, route planning, and intent inference  \cite{wang2025hybrid, wen2023dilu}. Extending this paradigm, VLMs integrate visual grounding with language-based reasoning, offering a multi-modal interface for interpreting complex scenes and human behaviors \cite{sima2024drivelm, xie2025vlms}.

Building upon this, the VLA framework seeks to unify scene understanding, interactive reasoning, and action generation within a single model \cite{zitkovich2023rt, kim2024openvla}. Typically, VLA approaches fine-tune pretrained VLMs on driving-related QA tasks or demonstration data, and leverages language-guided reasoning to bridge perception and control \cite{marcu2024lingoqa, kim2019grounding}. Unlike traditional perception and planning models, VLA enables more flexible, interpretable, and generalizable decision-making, especially in ambiguous or rare situations \cite{zhou2025opendrivevla, zhang2025safevla}. While still in its early stages, this line of research presents a promising foundation for building cognitively capable, human-aligned AV, offering key advantages in navigating uncertain and long-tail scenarios.


\section{Methodology}
\subsection{Overview}
\begin{figure*}[t]
  \begin{center}
  \centerline{\includegraphics[width=6.8in]{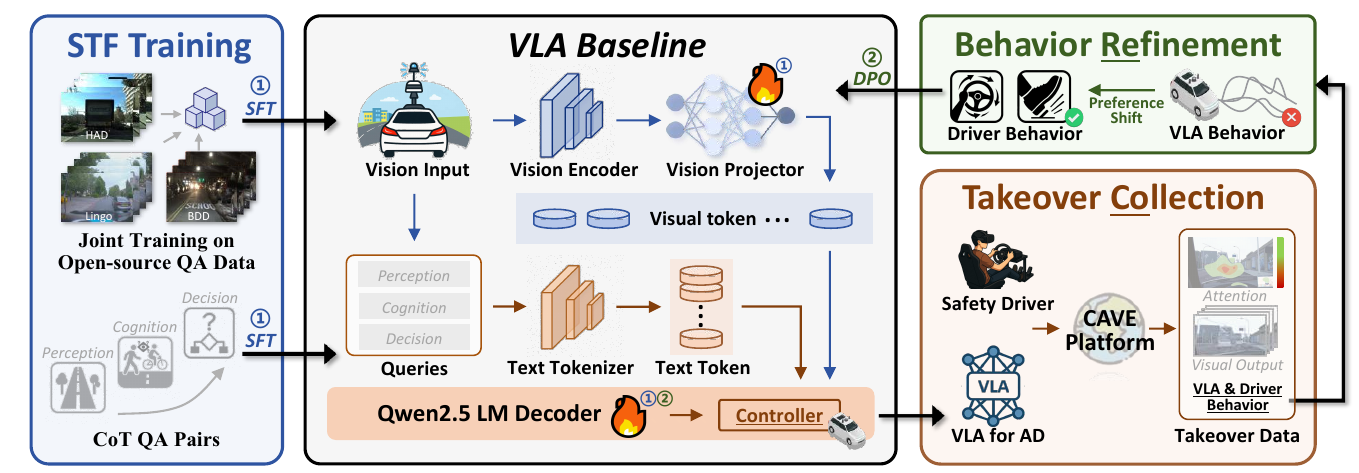}}
  \caption{Overview of the proposed CoReVLA framework.}\label{fig:framework}
  \end{center}
  \vspace{-0.6cm}
\end{figure*}

To improve AV performance in long-tail scenarios, we propose CoReVLA, as illustrated in Fig.~\ref{fig:framework}. First, the Qwen2.5-VL-7B model is STF with a combination of open-source driving QA datasets to build a foundational understanding of driving tasks. It is then deployed in the CAVE platform, a closed-loop, HITL simulation environment, where long-tail failure cases requiring human takeovers are identified and collected. Finally, CoReVLA is refined via DPO using human feedback from takeover events, enabling the model to align with human preference and improve its generalization in long-tail scenarios.

\subsection{Pre-Stage 1: SFT with QA Data}
\subsubsection{Data Construction.}


High-quality QA data is essential for enabling VLMs to comprehend domain-specific tasks and execute them effectively. To this end, we curate a 70GB domain-specific dataset by integrating LingoQA \cite{marcu2024lingoqa}, BDD \cite{kim2018textual}, and HAD \cite{kim2019grounding}. The dataset is organized into two parts: cognition for scenarios and action for learning safe driving strategies.

Each training instance includes five consecutive image frames at one-second intervals and structured QA pairs in a Chain-of-Thought (CoT) format. This design mirrors the human reasoning process from scene understanding to decision-making, enhancing both interpretability and behavioral soundness. 

\subsubsection{SFT Training.}

To adapt a general-purpose VLM to domain-specific reasoning tasks in autonomous driving, we performed supervised fine-tuning on the Qwen2.5-VL-7B model using the constructed dataset. Specifically, we applied Low-Rank Adaptation (LoRA) to two key components of the model: the vision projector and the LLM backbone. The former enhances the model’s ability to align visual inputs with textual semantics, while the latter improves its capacity to understand and reason about driving-related questions.

Specifically, the Qwen-VL architecture consists of a visual encoder, a vision-language projector, and a decoder-only transformer-based LLM. The visual encoder extracts patch-level features \( \mathbf{I} \in \mathbb{R}^{N \times D_v} \), where \( N \) is the number of image patches and \( D_v \) is the visual feature dimension. These features are projected into the LLM token embedding space \( \mathbf{V} \in \mathbb{R}^{N \times D_t} \) via a learned projection function \( f_{\text{proj}}: \mathbb{R}^{D_v} \rightarrow \mathbb{R}^{D_t} \), where \( D_t \) is the embedding size of the LLM. These tokens are then prepended or interleaved with text tokens and processed by the LLM for autoregressive generation.

To enable efficient domain adaptation, LoRA modules are inserted into selected linear layers of both the vision projector and the LLM transformer. Given a frozen pretrained weight \( \mathbf{W}_0 \in \mathbb{R}^{d \times k} \), where \( d\) and \( k\) denote the output and input dimensions of the linear layer respectively LoRA models the update as a low-rank decomposition \( \Delta\mathbf{W} = \mathbf{B} \mathbf{A} \), where \( \mathbf{A} \in \mathbb{R}^{r \times k} \) and \( \mathbf{B} \in \mathbb{R}^{d\times r} \) are the learnable LoRA parameters, and \( r \ll \min(d, k) \) is a user-defined rank hyperparameter controlling the adaptation capacity.

In addition, let the above dataset be denoted as \( \mathcal{D} = \{(a_i^{\text{img}}, a_i^{\text{text}}, b_i)\}_{i=1}^{N} \) consists of image sequences \( a_i^{\text{img}} \), textual prompts \( a_i^{\text{text}} \), and  target output sequences \( b_i \). The objective function is the standard autoregressive cross-entropy loss:

\begin{equation}
\mathcal{L}_{\text{SFT}} = - \sum_{i=1}^{N} \sum_{t=1}^{T} \log P_{\theta}(b_{i,t} \mid a_i^{\text{img}}, a_i^{\text{text}}, b_{i,<t})
\end{equation}
where \( b_{i,<t} = (b_{i,1}, \ldots, b_{i,t-1}) \), and \( \theta \) represents the set of trainable parameters introduced via the LoRA modules.

This fine-tuning strategy enables parameter-efficient and task-aligned adaptation of Qwen2.5-VL-7B, making it suitable for high-level visual reasoning tasks in autonomous driving scenarios. Detailed dataset information and SFT training parameters are provided in Appendix A.

\subsection{Stage 1: Takeover Data Collection}


\begin{figure*}[t]
  \begin{center}
  \centerline{\includegraphics[width=6.8in]{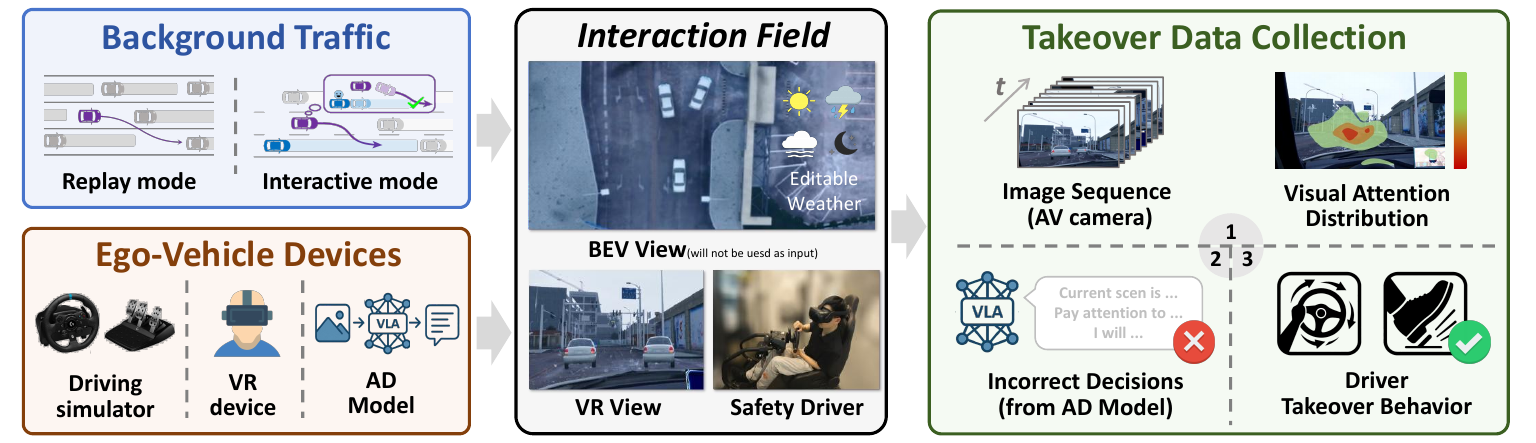}}
  \caption{Human-in-the-loop testing and takeover data collection in the CAVE platform.}\label{fig:CAVE}
  \end{center}
  \vspace{-0.6cm}
\end{figure*}

After SFT in Pre-Stage, the VLM gains a basic understanding of driving tasks and can perform reliably in routine scenarios under open-loop tests. However, real-world closed-loop driving introduces different challenges. Long-tail scenarios, while infrequent, are often responsible for the majority of safety-critical failures. Consequently, enhancing model performance under such rare but high-risk conditions remains an urgent priority. 

To collect driving data from long-tail scenarios, we consider human takeover events as representative failure cases that expose the limitations of the current model. Therefore, the intervention marks the boundary of the model’s capabilities and thus offers valuable guidance for enhancing robustness and safety.

To systematically collect such takeover data, we develop an immersive closed-loop testing platform named CAVE, as illustrated in Fig.~\ref{fig:CAVE}. The platform includes two main types of agents: the ego vehicle, controlled by the CoReVLA, and background traffic participants. The safety driver operates in a first-person perspective using a VR headset, enabling realistic visual feedback and timely intervention via a connected driving simulator. Background traffic can operate in either replay or interactive mode, allowing for targeted evaluation of perception, planning, and interaction performance across diverse scenarios.

In our experiments, CoReVLA is integrated into the CAVE platform, where it interacts with background vehicles in real time. Its performance is continuously monitored throughout each test case. When CoReVLA exhibits suboptimal behavior that leads to deadlock or collision, the system switches to replay mode. In this mode, a safety driver wears a VR headset to experience an immersive driving environment and closely supervises CoReVLA’s behavior. If a hazardous situation arises, the driver performs a manual takeover.

Each takeover instance is recorded as a structured data sample, consisting of the historical image inputs, the driver’s visual attention at the moment of takeover, the driver’s control actions, and the behavior generated by CoReVLA prior to the intervention. These samples are automatically processed into the DPO training format and incorporated into the training dataset as Stage 2 input, where CoReVLA’s behavior is further refined using DPO.



\subsection{Stage 2: Behavior Refinement with DPO}


In Stage 2, CoReVLA is refined using takeover data collected from the CAVE platform. Each sample consists of an action pair: the suboptimal behavior previously generated by the model and the corrective behavior performed by the safety driver in the same scenario. These comparisons encode implicit human preferences and serve as supervision for learning more desirable driving policies. To align the model with human intent, we adopt DPO, which fine-tunes the policy to favor actions consistent with human takeovers, thereby reducing repeated failures in similar high-risk situations.

Compared to other Reinforcement Learning from Human Feedback (RLHF) methods, such as PPO, DPO offers several advantages. It eliminates the need for an explicitly designed reward function, which is often difficult to define in complex long-tail scenarios. This avoids issues such as reward hacking and reduces reliance on manual reward engineering. Moreover, DPO can be trained directly on offline human demonstration data, substantially improving data efficiency. These properties make DPO particularly well-suited for learning from sparse long-tail events.

Specifically, we model the conditional policy distribution over actions given an observation \( x \) as:

\begin{equation}
\pi_\theta(y \mid x) = \frac{\exp(g_\theta(x, y))}{\sum_{y'} \exp(g_\theta(x, y'))}
\end{equation}
where \( y \) is a candidate action, and \( g_\theta(x, y) \) is a learned scoring function that reflects the model's preference over \( y \) in context \( x \) (e.g., a logit output from a language model). This formulation defines a differentiable implicit policy that can be optimized via gradient-based methods.

Given a pairwise preference tuple \( (x, y^+, y^-) \), where \( y^+ \) denotes the human-preferred action and \( y^- \) is the model-generated suboptimal action, we define the probability that the model prefers \( y^+ \) over \( y^- \) as:

\begin{equation}
P(y^+ \succ y^- \mid x) = \sigma\left( \beta \cdot \left( g_\theta(x, y^+) - g_\theta(x, y^-) \right) \right)
\end{equation}
where \( \sigma(\cdot) \) is the sigmoid function and \( \beta \) is a temperature hyperparameter controlling preference sharpness. The DPO objective minimizes the negative log-likelihood of human preferences:

\begin{equation}
\mathcal{L}_{\text{DPO}} = - \mathbb{E}_{(x, y^+, y^-) \sim \mathcal{D}} \left[ \log \sigma\left(P(y^+ \succ y^- \mid x)   \right) \right]
\end{equation}

Optimizing this loss encourages the model to assign higher scores to human-preferred actions, effectively aligning its policy with expert behavior in critical scenarios.

\begin{table*}[t]
    \caption{Comparison with other methods on open-loop QA evaluation over multiple datasets. Best results are \textbf{bolded}.}
    \centering
    \centering
    \begin{tabular}{c|c c c |c c c|c c c}
    \toprule
     \multirow{2}{*}{Models} & \multicolumn{3}{c}{Lingo} & \multicolumn{3}{c}{BDD} & \multicolumn{3}{c}{HAD}   \\
     \cline{2-10}
      & BLEU & R-1 & R-L & BLEU & R-1 & R-L & BLEU & R-1 & R-L\\
      \hline
      Qwen2.5-VL-7B \cite{bai2025qwen25vltechnicalreport} & 9.7 & 19.4 & 11.8 & 35.1 & 27.5 & 19.3 & 27.2 & 28.1 & 21.8\\
      Llava-7B \cite{liu2023llava} & 20.1 & 28.9 & 22.2 & 28.9 & 26.8 & 20.7 & 24.6 & 25.7 & 19.3 \\
      LlavaNext-7B \cite{liu2024llavanext} & 17.4 & 27.8 & 20.0 & 30.8 & 28.4 & 19.8 & 26.6 & 28.7 & 21.5\\
      Impromptu \cite{chi2025impromptu} & 24.8 & 34.1 & 28.3 & 30.6 & 29.9 & 19.5 & 25.5 & 32.4 & 25.0 \\
      \hline
      \cellcolor{blue!5}\textbf{CoReVLA (Ours)} & \cellcolor{blue!5}\textbf{66.8} & \cellcolor{blue!5}\textbf{74.7} & \cellcolor{blue!5}\textbf{70.7} & \cellcolor{blue!5}\textbf{45.8} & \cellcolor{blue!5}\textbf{37.6} & \cellcolor{blue!5}\textbf{30.0} & \cellcolor{blue!5}\textbf{30.2} & \cellcolor{blue!5}\textbf{39.1} & \cellcolor{blue!5}\textbf{33.0}\\ 
     \bottomrule
\end{tabular}

    \label{tab:qa}
\end{table*}

\begin{figure*}[t]
  \begin{center}
  \centerline{\includegraphics[width=6.5in]{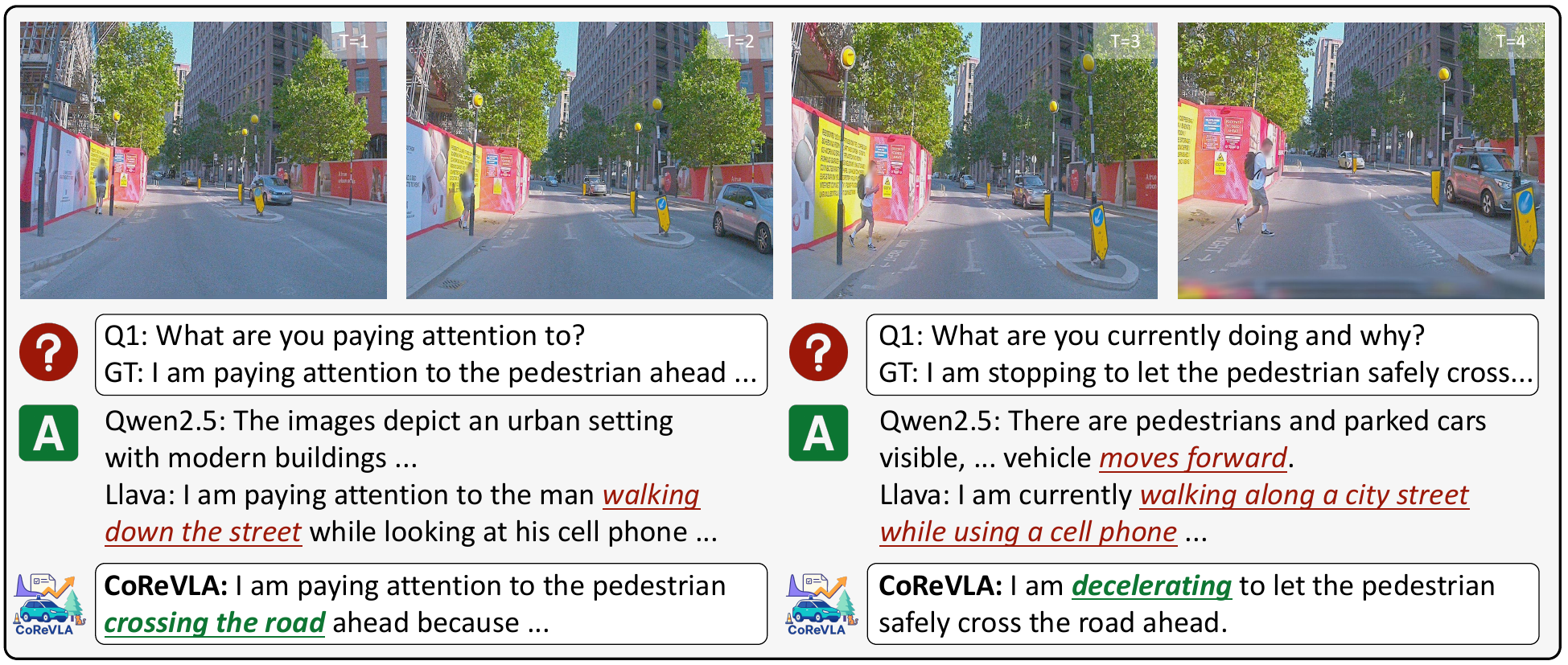}}
  \caption{Comparison of model responses in open-loop scene understanding and decision-making.}\label{fig:QA}
  \end{center}
  \vspace{-0.6cm}
\end{figure*}

To constrain policy drift and encourage stability during fine-tuning, some implementations further include a KL regularization term with respect to a reference policy \( \pi_{\text{ref}} \):

\begin{equation}
\mathcal{L}_{\text{total}} = \mathcal{L}_{\text{DPO}} + \lambda \cdot \text{KL}(\pi_\theta \| \pi_{\text{ref}})
\end{equation}
where \( \lambda \) is a regularization coefficient that controls the trade-off between preference alignment and proximity to the original policy. This regularization helps prevent overfitting to limited preference data while retaining generalization capabilities. Detailed descriptions of the DPO training dataset are provided in Appendix B.2.

In summary, DPO fine-tuning enables the model to capture human decision preferences in sparse data, significantly enhancing its generalization and safety performance in long-tail scenarios. Furthermore, by iteratively combining CAVE-based closed-loop testing with behavior refinement, the proposed CoReVLA can continuously evolve through a cycle of deployment, feedback, and adaptation—ultimately helping the model avoid repeated failures in similar scenarios.

\section{Experiment}
To evaluate whether CoReVLA can understand complex scenarios and complete driving tasks, we conduct both open-loop and closed-loop experiments. First, we compare its performance with baselines using BLEU and ROUGE. Then, we integrate CoReVLA into the CAVE platform to identify failure cases and apply DPO for behavior refinement. Finally, we benchmark against SOTA methods under closed-loop settings using Bench2Drive, which consists of diverse and challenging long-tail scenarios.

\subsection{Open-loop QA Evaluation}
To assess the language understanding and reasoning capability of CoReVLA, we first conduct open-loop QA evaluations across three representative datasets: LingoQA, BDD, and HAD. As shown in Tab.~\ref{tab:qa}, CoReVLA consistently achieves higher BLEU and ROUGE scores across all datasets, indicating that SFT enhances the model’s ability to understand driving scenarios and make correct decisions, laying the groundwork for closed-loop evaluation.

Fig.~\ref{fig:QA} further compares responses of different baseline models on scene understanding and decision-making. For the perception task, Llava correctly identified the pedestrian on the left as the most salient object, but failed to predict the pedestrian’s motion accurately. Regarding the decision-making task, both Qwen2.5 and Llava were unable to generate appropriate driving actions, resulting in potentially hazardous outcomes. In contrast, the proposed CoReVLA accurately inferred the intent of surrounding traffic participants and produced context-aware, safe driving decisions.

While CoReVLA achieves strong performance across open-loop QA benchmarks, such results are expected due to extensive QA-based pretraining in the Pre-Stage, which enhances the model’s language understanding and reasoning over routine driving scenarios.

However, excelling in static QA problems does not guarantee reliable behavior in real-world driving situations, especially under long-tail scenarios with high-risk. To more thoroughly assess CoReVLA’s capability under dynamic conditions, we conduct closed-loop driving evaluations, examining whether the model can make safe decisions and improve continually through human feedback.

\begin{figure*}[htbp]
  \begin{center}
  \centerline{\includegraphics[width=6.5in]{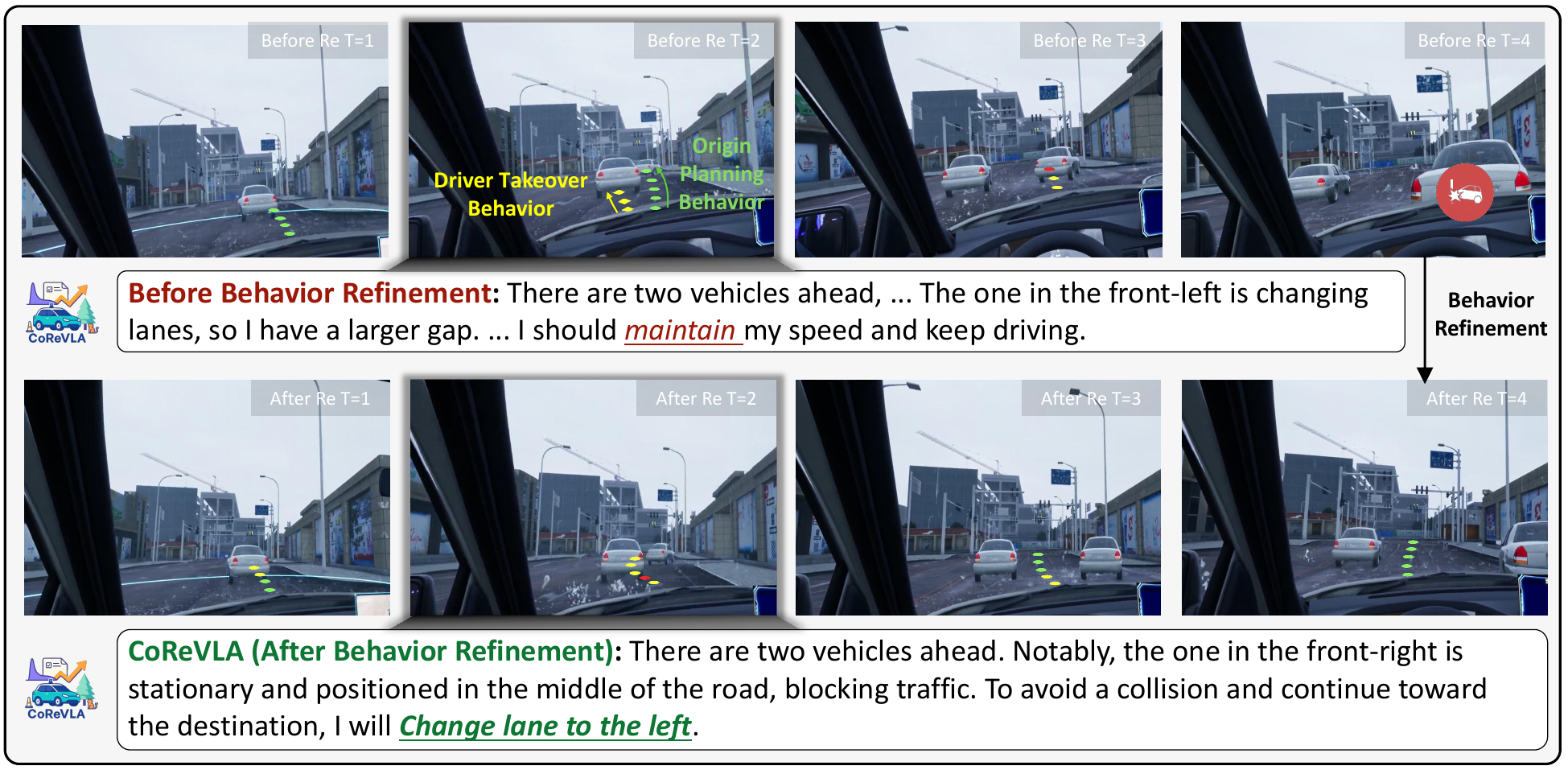}}
  \caption{Comparison of CoReVLA driving trajectories before and after behavior refinement.}\label{fig:takeover}
  \end{center}
  \vspace{-0.6cm}
\end{figure*}

\begin{table*}[ht]
\caption{Comparison of SOTA methods under closed-loop testing on the Bench2Drive benchmark. Best results are \textbf{bolded}.}
\centering
\small
\begin{tabular}{l|l|cccc}
\toprule
\textbf{Type} & \textbf{Method}  & \textbf{DS↑} & \textbf{SR(\%)↑} & \textbf{Efficiency↑} & \textbf{Comfortness↑}  \\
\midrule
\multirow{7}{*}{\makecell{Small-scale \\task-specific models}}& AD-MLP \cite{zhai2023rethinking} & 18.05 & 0.00 & 48.45 & 22.63  \\
& TCP \cite{wu2022trajectory} & 40.70 & 15.00 & 54.26 & \textbf{47.80}  \\
& VAD \cite{jiang2023vad} & 42.35 & 15.00 & \textbf{157.94} & 46.01 \\
& UniAD-Base \cite{hu2023planning}  & 45.81 & 16.36 & 129.21 & 43.58  \\
& ThinkTwice \cite{jia2023think} & 62.44 & 31.23 & 69.33 & 16.22  \\
& DriveAdapter \cite{jia2023driveadapter} & 64.22 & 33.08 & 70.22 & 16.01  \\
& DriveTransformer-Large \cite{jia2025drivetransformer} &  63.46 & 35.01 & 100.64 & 20.78  \\
\midrule
\multirow{4}{*}{\makecell{Large-scale \\ pretrained models}} & Impromptu* \cite{chi2025impromptu} & 19.38 & 0.00 & 37.96 & 39.31  \\
& InternVL* \cite{zhu2025internvl3} &  33.78 & 0.00 & 150.68 & 41.78  \\
& CoReVLA (Before refinement)* & 53.26 & 20.00 & 91.14 & 19.34  \\
\rowcolor{blue!5}
& \cellcolor{blue!5}\textbf{CoReVLA*} & \cellcolor{blue!5}\textbf{72.18}{\scriptsize (+7.96)} & \cellcolor{blue!5}\textbf{50.00}{\scriptsize (+14.99)} & \cellcolor{blue!5}145.41 & \cellcolor{blue!5}34.35 \\
\bottomrule
\end{tabular}
\label{tab:comparison}
\end{table*}

\subsection{Closed-loop Driving Evaluation}



Our closed-loop evaluation consists of two parts. In the first part, the Pre-Stage fine-tuned model is evaluated in the CAVE simulation platform under complex scenarios, with a HITL refinement process where a safety driver intervenes in failure cases to correct its behavior. Then, the refined model, CoReVLA, is integrated into the Bench2Drive benchmark for performance comparison against SOTA methods.

To recreate high-risk scenarios, we embed reconstructed 2D trajectory data into interactive background traffic within CAVE. Failure cases are logged, replayed, and used for human-driven takeover refinements. These refinements are then used to fine-tune the model further. Fig.~\ref{fig:takeover} shows how CoReVLA behaves before and after refinement. The color of each trajectory point indicates the vehicle's speed, with warmer colors representing lower velocities.

As illustrated in Fig.~\ref{fig:takeover}, this scenario unfolds on a rainy day, where the ego vehicle is following another car that suddenly changes lanes, exposing a stationary broken-down vehicle ahead. Before refinement, CoReVLA misinterpreted the lane change as an opportunity for increased driving space and chose to maintain its speed. It only began to react to the stationary vehicle moments later, initiating emergency braking too late to prevent a collision. Therefore, this critical scenario was extracted and replayed in the CAVE, during which a human driver intervened. Notably, the driver's attention was more focused on the stationary vehicle in the right front than on the lane-changing car on the left. The diamond-shaped trajectory in the figure represents the path taken after the human takeover. By using the takeover behavior, the pre-refinement model action, and the corresponding visual input as DPO training data, the model is fine-tuned to better align with human intent. After refinement, CoReVLA was able to recognize the potential risk earlier and proactively execute a lane change, successfully avoiding the collision.


\begin{figure*}[htbp]
  \begin{center}
  \centerline{\includegraphics[width=6.5in]{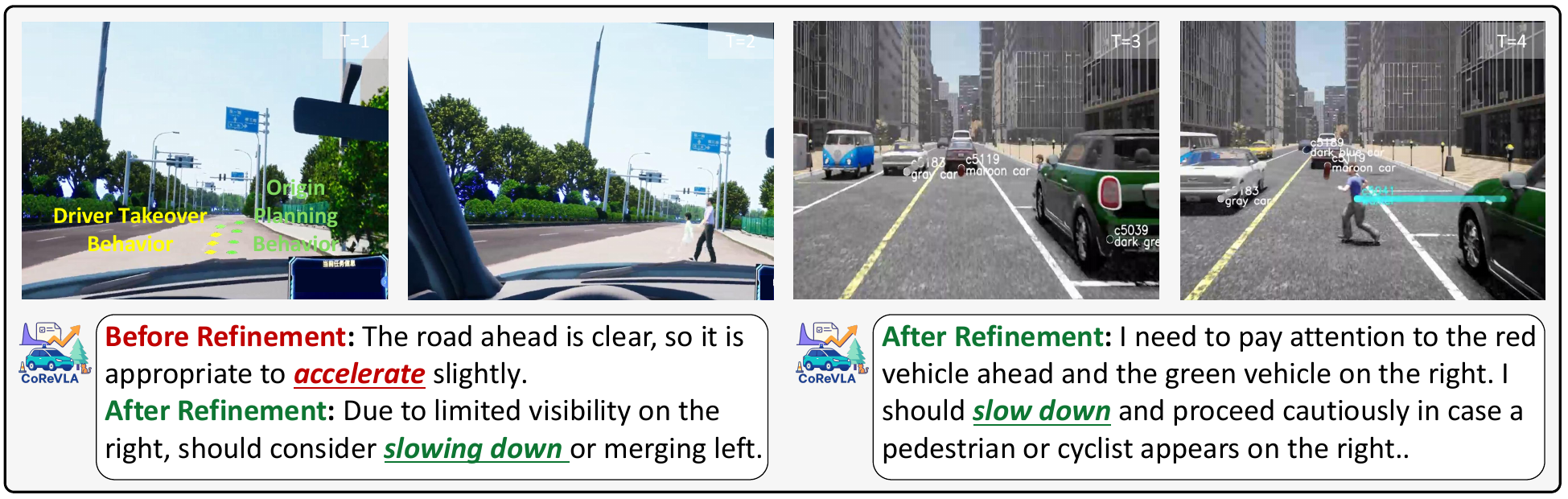}}
  \caption{Scenario comparison illustrating CoReVLA’s behavior refinement and generalization across CAVE and Bench2Drive.}\label{fig:carla}
  \end{center}
  \vspace{-0.6cm}
\end{figure*}

After completing the behavior refinement within the CAVE platform, we further evaluate the resulting CoReVLA model using the Bench2Drive benchmark. We compare the proposed CoReVLA against SOTA end-to-end autonomous driving methods across multiple metrics, including DS and SR. Methods marked with an asterisk are evaluated on the Dev10 dataset introduced by DriveTransformer \cite{jia2025drivetransformer}. Detailed scenario descriptions can be found in Appendix C.1.

Tab.~\ref{tab:comparison} presents the performance of several representative methods from both small-scale task-specific models and large-scale pretrained models on the Bench2Drive benchmark. Compared to existing SOTA approaches, our proposed CoReVLA achieves the highest DS and SR, reaching 72.18 and 50.00\%, respectively. This corresponds to an improvement of 7.96 points in DS and a 14.99\% increase in SR over the second-best method.

While CoReVLA demonstrates significant improvements in DS and SR, it does not outperform all baseline models in terms of efficiency and comfortness. This is mainly because CoReVLA focuses on high-risk, long-tail driving scenarios where safety is prioritized during model refinement. In the DPO-based HITL fine-tuning within the CAVE platform, drivers tend to exhibit cautious behavior, maintaining moderate speeds and carefully observing their surroundings, rather than accelerating quickly to exit potentially dangerous situations. Additionally, emergency braking is sometimes required for safety, which can negatively impact comfort-related metrics. This explains why, despite a significant increase in SR, the improvement in DS is relatively modest. A similar pattern is observed in DriveTransformer-Large, which is the second-best performing model.

In addition, as shown in Tab.~\ref{tab:comparison}, both DS and SR improve significantly after behavior refinement, further demonstrating the effectiveness of the proposed \textbf{Collect-and-Refine} dual-stage process in enhancing AD performance in long-tail scenarios.


Finally, Fig.~\ref{fig:carla} shows a CAVE-constructed scenario that mirrors a similar case in the Bench2Drive dataset, enabling a direct comparison of model generalization across platforms. This case demonstrates that behavior refinement based on human takeover data in CAVE can effectively generalize to similar scenarios. This evidence shows that CoReVLA is capable of continual learning and behavioral evolution, avoiding repeated failures in comparable scenarios.

Specifically, in the CAVE platform, we constructed a scenario where a pedestrian suddenly emerges from roadside vegetation. Before refinement, the model perceives no immediate obstacles on the road and thus accelerates toward a high desired speed. Therefore, when the pedestrian appears, the model is unable to react in time, resulting in a collision. In contrast, during the HITL replay, the driver slows down upon entering an area with heavy roadside occlusion and begins monitoring the roadside area. This proactive approach allows the driver to spot the pedestrian early and brake safely, thus avoiding the collision.

To evaluate the generalization ability of the proposed CoReVLA, we selected scenario \#3255 from the Bench2Drive benchmark, where a pedestrian unexpectedly runs into the road in front of a stopped green vehicle. As illustrated in the figure, the refined CoReVLA successfully transfers the learned behavior by reducing speed in areas with limited roadside visibility and ensuring it can brake in time to complete the scenario.

\section{Conclusion}
Current autonomous driving systems continue to underperform in long-tail, safety-critical scenarios, primarily due to the scarcity of high-value QA data and the lack of efficient training strategies.
To address this, we propose CoReVLA, a continual end-to-end driving framework with a dual-stage Collect-and-Refine process.
By testing the model in the CAVE platform and collecting driver takeover data, CoReVLA leverages DPO to refine its behavior in alignment with human preferences. 
We validate CoReVLA through both open-loop and closed-loop experiments. Open-loop QA evaluations across three open-source datasets demonstrate substantial improvements in language understanding and decision-making capabilities. In closed-loop tests on the Bench2Drive benchmark, CoReVLA achieves a DS of 72.18 and a SR of 50.00\%, surpassing the best prior method by 7.96 DS and 15.00\% SR. Case studies further confirm CoReVLA’s ability to continually refine its policy and generalize to similar failure-prone scenarios by learning from past human takeovers.
In summary, this work establishes a complete pipeline from HITL data collection to behavior refinement, offering a practical paradigm for improving AD in long-tail scenarios. Future research will explore real-world deployment and incorporate richer forms of human feedback.

\ifCLASSOPTIONcaptionsoff
  \newpage
\fi

\footnotesize
\bibliographystyle{IEEEtranN}
\bibliography{IEEEabrv,Bibliography}

\vfill

\end{document}